\documentclass[11pt]{article}

\usepackage[utf8]{inputenc}
\usepackage[T1]{fontenc}
\usepackage{amsmath,amssymb}
\usepackage{booktabs}
\usepackage{graphicx}
\usepackage{microtype}
\usepackage[margin=1in]{geometry}
\usepackage[round]{natbib}
\usepackage[hidelinks]{hyperref}
\usepackage{xcolor}

\title{Depth-Staggered Fibonacci Spacing for Sparse Attention:\\
Static Schedules Beat Learned Dilation and\\
Extrapolate Where Dense Attention Fails\thanks{Code, the paper source, and the
experiment database are available at
\url{https://github.com/ccapps42/scaled-fibonacci-attention}.}}

\author{%
  Chad A. Capps\\
  Independent Researcher\\
  \texttt{Chad.Capps@me.com}
}
\date{\today}

\begin{document}
\maketitle

\begin{abstract}
We study sparse self-attention in which each query attends to a dense local
window plus a set of Fibonacci-spaced offsets, with a per-layer scalar $\alpha$
that compresses or expands the spacing. Across 21 language models trained under
one matched recipe (60M parameters, 512 hidden, 16 layers, 426M tokens), we
compare four ways of setting $\alpha$ across depth: fixed, per-layer learned,
a static linear stagger, and a coprime (anti-gridding) reassignment of that
stagger, together with a reach-matched power-of-2 control. Three results stand
out. First, a static per-layer stagger improves perplexity over both fixed and
learned $\alpha$, and the gain is base-agnostic: applying the same stagger to a
power-of-2 base lifts it above fixed Fibonacci and to parity with learned
Fibonacci attention. Second,
learning $\alpha$ per layer is inert: it does not beat the static schedule and
costs roughly five times the inference latency. Third, and most consequential,
all sparse variants extrapolate to four times their training length with little
or no degradation, whereas a recipe-matched dense baseline collapses
(perplexity rises by 201\% at 4$\times$ length); we attribute this to
fixed-offset attention only ever querying relative positions seen during
training. We also report two honest negatives: at training length the best sparse
model has about 26\% higher perplexity than the dense baseline, and the
staggering gain is uniform across context positions rather than concentrated at
long range.
\end{abstract}

\section{Introduction}

Sparse attention reduces the quadratic cost of self-attention by restricting
each query to a subset of keys. A recurring design question is \emph{which}
subset, and in particular how the chosen offsets should vary with depth. Many
schemes fix one pattern and rely on composition through the stack to widen the
receptive field; others vary the pattern across heads. Comparatively little is
known about varying a single spacing parameter \emph{per layer} in a controlled
language-modeling setting.

We take the offsets to follow a Fibonacci sequence (denser at short range,
geometrically sparser at long range) and attach a per-layer scalar $\alpha$
that scales every offset, $d_k = \alpha f_k$. The original motivation was a
``spring'': let each layer \emph{learn} its own $\alpha$ so that depth-appropriate
spacing emerges during training. That hypothesis is not supported by our data.
The learned scalar barely moves from its initialization and yields no advantage
over a fixed schedule. What does help is setting $\alpha$ \emph{statically} but
differently per layer, a fixed stagger across depth that we imposed rather than
learned. This paper reports the full arc, including the negative result on
learning, because the static schedule that wins is simpler than the mechanism we
set out to study.

Our contributions are:
\begin{itemize}
\item A controlled comparison of four per-layer $\alpha$ schedules (fixed,
learned, static linear stagger, coprime stagger) and a reach-matched power-of-2
control, all sharing one attention implementation and one training recipe, over
21 trained models and a window sweep $W \in \{6,8,10,12\}$
(Section~\ref{sec:results}).
\item Evidence that a static per-layer stagger is the dominant lever: it beats
both fixed and learned $\alpha$, and it is \emph{base-agnostic}: staggering a
power-of-2 base raises it above fixed Fibonacci and to parity with learned
Fibonacci attention. The Fibonacci base then adds a separable gain on top
(Section~\ref{sec:ablation}).
\item A length-extrapolation result: structured sparse attention is robust out
to 4$\times$ training length while a recipe-matched dense baseline collapses,
with a mechanism (no out-of-distribution relative positions) that also explains
the ordering among sparse variants (Section~\ref{sec:extrap}).
\item Two honest negatives that bound the claims: a training-length quality gap
to dense, and a position-resolved analysis showing the staggering gain is
uniform rather than long-range-specific (Section~\ref{sec:negatives}).
\end{itemize}

\section{Method}
\label{sec:method}

\paragraph{Attended set.}
For a query at position $i$, the attended key set is the union of a dense local
window and a set of scaled Fibonacci offsets. The window covers distances
$\{0,1,\dots,W\}$ (self-position included, so no query has an empty key set).
The sparse offsets use the base sequence
\[
\mathrm{fib} = [1,2,3,5,8,13,21,34,55,89,144,233,377,610,987],\qquad K=15,
\]
with each rung scaled to a target distance $d_k = \alpha f_k$ and gathered at the
key/value rows $i - d_k$. Because $d_k$ is continuous, we gather by linear
interpolation between the floor and floor-plus-one positions, which keeps a
nonzero gradient to $\alpha$ even when every rung coincides with an integer position. Rotary
position embeddings are applied at true positions before gathering. Offsets are
shared across all queries, so this is a set of shifted gathers rather than
per-position index lists.

\paragraph{The per-layer scalar.}
We write $\alpha = 0.5 + \mathrm{sigmoid}(\theta) \in [0.5, 1.5]$, with $\theta$
the only added parameter. The base sequence already spans the context at
$\alpha=1$ ($f_{15}=987 \approx 1024$), so the useful dynamic range is
compression ($\alpha<1$) plus a small margin above 1. We study four ways of
setting $\theta$ across the $L=16$ layers:
\begin{itemize}
\item \textbf{Fixed:} $\alpha_\ell = 1$ for all layers.
\item \textbf{Learned:} $\theta_\ell$ is a free parameter per layer, initialized
to $\alpha=1$.
\item \textbf{Staggered (linear):} $\alpha_\ell = 0.5 + \ell/(L-1)$, a fixed ramp
from $0.5$ at layer $0$ to $1.5$ at layer $L-1$.
\item \textbf{Staggered (coprime / HDC):} the same multiset of $\alpha$ values,
reassigned to layers by a coprime stride $\ell \mapsto (S\ell \bmod L)$ with
$S=7$, so adjacent layers receive maximally different spacing. This ports the
anti-gridding idea from dilated convolutions \citep{wang2018understanding} to
per-layer attention spacing.
\end{itemize}

\paragraph{Control base.}
As a control we replace the Fibonacci base with a reach-matched power-of-2 base
(log-sparse). It reaches the same maximum distance with fewer rungs, isolating
the effect of spacing \emph{density} at equal reach rather than rung count.

\paragraph{Implementation note.}
A hard boolean attention mask carries no gradient to $\alpha$, so the learned
variant uses the interpolated gather. Fixed and staggered variants have integer
offsets and could use a fused boolean-mask path, but we report them through the
same gather module to remove any numerical-path confound; for the
length-extrapolation study the integer-offset variants additionally run through a
fused boolean-mask attention at evaluation time, which is numerically identical
and faster. All variants share projections, GQA shapes, and RoPE with the dense
baseline; the only added parameters are the $\theta$ scalars.

\section{Experimental Setup}
\label{sec:setup}

\paragraph{Model and training.}
All models use the same architecture: $d_{\text{model}}=512$, 16 layers, 8
attention heads, 2 key/value heads (GQA), head dimension 64, vocabulary 32{,}768,
sequence length 1024, SwiGLU feed-forward with ratio $8/3$, RMSNorm, RoPE
($\theta_{\text{rope}}=10^4$), tied embeddings, no dropout. Training is identical
across runs: 13{,}000 steps at an effective batch of 32{,}768 tokens
(426M tokens total), AdamW ($\beta=0.9,0.95$, weight decay $0.1$), learning rate
$3\times10^{-4}$ with 300 warmup steps and cosine decay, gradient clip $1.0$,
bf16 autocast, seed 42. The dense baseline (denoted \textsc{dense}) shares this
recipe exactly; the only difference is its attention mechanism (verified by a
field-by-field config comparison). Total non-embedding parameters are 44.1M
(60.9M including the tied embedding).

\paragraph{Data and evaluation.}
Models train on a fixed token mixture, in approximate token proportions $35\%$
FineWeb-Edu \citep{penedo2024fineweb}, $22\%$ Wikipedia \citep{wikidump}, $20\%$
TinyStories \citep{eldan2023tinystories}, and $23\%$ mathematics (the MATH dataset
\citep{hendrycks2021math} and OpenMathInstruct-2
\citep{toshniwal2024openmathinstruct}). We evaluate by token-level perplexity on
four held-out sets: FineWeb-Edu, Wikipedia, TinyStories, and a math set held out
from OpenMathInstruct-2 (problem--solution pairs, a disjoint shard from the
training math).
We additionally log a battery of auxiliary evaluations (two synthetic multi-hop
reasoning tasks, RULER variable tracking \citep{hsieh2024ruler} and LEGO
\citep{zhang2022lego}, plus a set of cheap probes carried over from the training
harness) and an analytic FLOPs/throughput breakdown. At this scale the battery is
uniformly capacity-limited (Section~\ref{sec:negatives}). The sparse models add only the $\theta$
scalars over dense, so the comparison to \textsc{dense} is essentially
parameter-matched.

\paragraph{Run matrix.}
We sweep the local window $W \in \{6,8,10,12\}$ for each of: fixed Fibonacci,
learned Fibonacci, linear-staggered Fibonacci, coprime-staggered (HDC)
Fibonacci, and linear-staggered power-of-2. The plain power-of-2 control (fixed,
$\alpha=1$) is run at $W=8$. This yields 21 trained models. Throughput note: the
learned variant runs through the unfused interpolated gather and is roughly
$5\times$ slower at inference than the fused integer-offset variants at identical
FLOPs; we treat efficiency as context, not as a selection criterion, since at
this shape the attention term is a minority of per-layer compute.

\section{Results}
\label{sec:results}

\subsection{Perplexity: staggering is the dominant lever}

Table~\ref{tab:matrix} reports held-out perplexity for the full matrix. Within
the Fibonacci base, the ordering is consistent across every window and dataset:
\[
\text{staggered} \approx \text{coprime} \;>\; \text{learned} \;>\; \text{fixed},
\]
and the best model overall is the linear stagger at $W=12$ (FineWeb 42.02,
Wikipedia 37.59). The coprime reassignment ties the linear stagger to within
seed-scale differences at every window, so the \emph{cascade order} of the
per-layer spacings does not matter; only the multiset of spacings does. The
plain power-of-2 control is the weakest model in the matrix.

\begin{table}[htbp]
\centering
\small
\begin{tabular}{llrrrr}
\toprule
base / schedule & $W$ & TinyStories & Math & FineWeb-Edu & Wikipedia \\
\midrule
fib / fixed     & 6  & 5.57 & 3.45 & 44.35 & 40.24 \\
fib / fixed     & 8  & 5.55 & 3.45 & 44.08 & 40.02 \\
fib / fixed     & 10 & 5.48 & 3.42 & 43.48 & 39.33 \\
fib / fixed     & 12 & 5.44 & 3.39 & 43.12 & 39.05 \\
\addlinespace
fib / learned   & 6  & 5.53 & 3.40 & 43.86 & 39.62 \\
fib / learned   & 8  & 5.52 & 3.40 & 43.70 & 39.43 \\
fib / learned   & 10 & 5.43 & 3.35 & 42.99 & 38.80 \\
fib / learned   & 12 & 5.40 & 3.35 & 42.85 & 38.60 \\
\addlinespace
fib / stagger   & 6  & 5.39 & 3.28 & 42.93 & 38.59 \\
fib / stagger   & 8  & 5.36 & 3.27 & 42.53 & 38.07 \\
fib / stagger   & 10 & 5.33 & 3.26 & 42.21 & 37.88 \\
\textbf{fib / stagger} & \textbf{12} & \textbf{5.30} & \textbf{3.25} & \textbf{42.02} & \textbf{37.59} \\
\addlinespace
fib / coprime   & 6  & 5.40 & 3.29 & 43.10 & 38.70 \\
fib / coprime   & 8  & 5.37 & 3.28 & 42.58 & 38.27 \\
fib / coprime   & 10 & 5.33 & 3.27 & 42.28 & 38.00 \\
fib / coprime   & 12 & 5.30 & 3.26 & 42.09 & 37.75 \\
\addlinespace
pow2 / fixed    & 8  & 5.67 & 3.55 & 44.78 & 40.89 \\
pow2 / stagger  & 6  & 5.54 & 3.43 & 44.11 & 39.89 \\
pow2 / stagger  & 8  & 5.48 & 3.41 & 43.60 & 39.56 \\
pow2 / stagger  & 10 & 5.43 & 3.39 & 43.17 & 39.05 \\
pow2 / stagger  & 12 & 5.38 & 3.36 & 42.69 & 38.51 \\
\bottomrule
\end{tabular}
\caption{Held-out perplexity for all 21 models (lower is better). Within the
Fibonacci base, staggered $\approx$ coprime $>$ learned $>$ fixed at every
window. The best model is the linear stagger at $W{=}12$.}
\label{tab:matrix}
\end{table}

\subsection{The staggering gain is base-agnostic, and the base adds to it}
\label{sec:ablation}

Staggering the power-of-2 base lifts it substantially: at $W=8$, plain
log-sparse improves from 44.78 to 43.60 FineWeb perplexity once the same linear
$\alpha$ ramp is applied. The cleanest statement of the effect uses the
three-seed means at $W{=}12$ (Table~\ref{tab:seed}): \emph{power-of-2 staggered
($42.84$) beats fixed Fibonacci ($43.18$) and matches learned Fibonacci
($42.85$)}. A worse base, merely staggered, equals or outperforms the better base
whether it is fixed or learned. So per-layer spacing diversity dominates the base
choice, and learning $\alpha$ buys nothing that staggering does not. (A single
seed had suggested power-of-2 staggered edged ahead of learned Fibonacci by
$0.16$; across three seeds that gap vanishes to $0.01$, within the seed-noise
band of Table~\ref{tab:seed}, so we report it as a match.)

The Fibonacci base nonetheless contributes a separable gain. At matched
staggering, Fibonacci beats power-of-2 at every window (FineWeb: 42.93 vs.\ 44.11
at $W{=}6$ down to 42.02 vs.\ 42.69 at $W{=}12$). The two factors (staggering
and base density) are additive rather than redundant.

\paragraph{Why staggering widens coverage.}
At fixed $\alpha=1$ every layer attends the same offsets, so the set of distinct
distances reachable anywhere in the stack equals one layer's offsets (about 19
distinct distances beyond the window at sequence length 1024). A linear $\alpha$
ramp gives each layer a different scaled copy of the same rungs, raising the
union of reachable distances by roughly $4\times$ (to about 75) at the same
per-layer sparsity and the same attention FLOPs.

This count of distinct distances still understates the receptive field, because
the rungs are not point samples. The dense local window of width $W$ acts as a
smoothing kernel at every layer: within a few layers each token's representation
already aggregates its $\pm W$ neighbors, and the blur compounds with depth. A
rung read at distance $d_k=\alpha f_k$ therefore returns a soft band of half-width
on the order of $W$ centered on $d_k$, not a single position. Consecutive
Fibonacci rungs are separated by $f_{k+1}-f_k=f_{k-1}$, so whenever that spacing
falls below about $2W$ the bands of neighboring rungs overlap and the apparent gap
between them is not a real hole. This is automatic at short range, where the rungs
are dense ($1,2,3,5,8,\dots$); the spacing only exceeds $2W$ far out (for example
$610\to987$), where the precise offset matters least. Staggering reinforces the
effect: because each layer scales the same nominal rung to a different absolute
distance, the union over layers places several bands in an interval that any
single layer would leave empty, and multi-hop composition adds the sums
$d_a+d_b$ of rungs used in successive layers. The distinct-distance count is thus
a conservative lower bound on coverage.

This argument explains why the scheme has no functional gaps, not where its
measured advantage comes from. The position-resolved analysis in
Section~\ref{sec:negatives} shows the staggering gain is uniform across context
positions rather than concentrated at long range, so the widened receptive field
is better read as evidence that staggering sacrifices no coverage than as the
source of the perplexity improvement.

\subsection{Learning the scalar is inert}
\label{sec:inert}

The learned schedule improves over fixed (Table~\ref{tab:matrix}) but never
reaches the static stagger, and it does so at roughly $5\times$ the inference
latency because the non-integer offsets require the unfused gather. The learned
$\alpha$ values stay close to their initialization, which a controlled retrieval
probe (Section~\ref{sec:probe}) confirms directly. The mechanism we set out to
study, a trainable spring, is therefore dominated by a fixed schedule we
imposed by hand. We read this as evidence that the benefit is structural (diverse
per-layer spacing) rather than something gradient descent needs to discover.

\subsection{Mechanistic probe: the scalar does not move}
\label{sec:probe}

To test the inertness directly, away from language modeling, we use a controlled
retrieval task. Each sequence plants a key--value pair at a chosen offset $d$ from
the query, with distractor pairs elsewhere so the model must perform genuine
distance-dependent retrieval rather than copy the only value present. A marked
variant asks a single model to serve two distances at once ($d\in\{200,350\}$), so
the per-layer scalars would have to specialize to cover both. We train small models
at depths $2$, $4$, and $8$ with the spacing scalar free to learn, under two
initializations: uniform ($\alpha=1$ in every layer) and the staggered ramp.
Table~\ref{tab:probe} reports the learned scalars and the retrieval accuracy.

Two results follow. First, the free scalar barely moves from its initialization at
any depth: the uniform start stays within $0.05$ of $\alpha=1$, and the staggered
start stays spread across roughly $[0.55,1.45]$. Gradient descent does not
relocate the scalars even on a task that would directly reward it, which confirms
the language-model observation of Section~\ref{sec:inert} in a setting where the
useful spacing is unambiguous. Second, the initialization decides whether the
model retrieves at all: with the uniform start it retrieves nothing at depth $4$
and beyond ($0.00$ at both distances), while the staggered start restores retrieval
at depths 2 and 4.

We take this as direct support for two claims of this paper: the per-layer scalar
is inert under learning, and the static staggered \emph{structure}, not adaptation,
is what carries the benefit. The probe does not support a stronger union-coverage
claim. Even when it retrieves, the staggered model tends to specialize to one
distance (high at $200$, near zero at $350$), and at depth $8$ the task is
unsolved. This synthetic task isolates the inertness cleanly but demands more of
coverage than language modeling does; the coverage argument therefore rests on the
distinct-distance count of Section~\ref{sec:ablation}, not on this probe.

\begin{table}[htbp]
\centering
\small
\begin{tabular}{llccc}
\toprule
depth & init & learned $\alpha$ range & acc@$200$ & acc@$350$ \\
\midrule
2 & uniform   & $0.93$--$1.03$ & 0.38 & 0.23 \\
2 & staggered & $0.56$--$1.39$ & 0.99 & 0.00 \\
\addlinespace
4 & uniform   & $0.96$--$1.01$ & 0.00 & 0.00 \\
4 & staggered & $0.55$--$1.45$ & 0.88 & 0.17 \\
\addlinespace
8 & uniform   & $0.97$--$1.02$ & 0.00 & 0.00 \\
8 & staggered & $0.55$--$1.45$ & 0.04 & 0.00 \\
\bottomrule
\end{tabular}
\caption{Controlled retrieval probe (marked two-distance copy, $d\in\{200,350\}$),
mean over two seeds. A spacing scalar left free to learn stays near its
initialization at every depth. The uniform start collapses to no retrieval at depth
$\ge 4$; the staggered start restores it at depths 2 and 4, though it specializes
to one distance and the task is unsolved at depth 8.}
\label{tab:probe}
\end{table}

\subsection{Length extrapolation: sparse is robust, dense collapses}
\label{sec:extrap}

All models train at sequence length 1024. We rebuild each model at evaluation
lengths 2048 and 4096 (RoPE and the offset set extend naturally; no
length-dependent parameters exist) and measure perplexity.
Table~\ref{tab:extrap} shows the result. The structured sparse models are flat to
slightly improving out to 4$\times$ length. The dense baseline, best at training
length, degrades catastrophically: FineWeb perplexity rises from 33.37 to 100.36,
a 201\% increase, and is already behind the sparse models by 2048.

\begin{table}[htbp]
\centering
\small
\begin{tabular}{lrrrr}
\toprule
model & $T{=}1024$ & $T{=}2048$ & $T{=}4096$ & change at $4\times$ \\
\midrule
\textsc{dense}        & \textbf{33.37} & 49.15 & 100.36 & $+201\%$ \\
fib / stagger $W{=}12$ & 42.02 & 41.65 & 41.45 & $-1.3\%$ \\
fib / stagger $W{=}8$  & 42.53 & 42.17 & 41.98 & $-1.3\%$ \\
fib / coprime $W{=}12$ & 42.09 & 42.53 & 43.25 & $+2.8\%$ \\
fib / fixed $W{=}8$    & 44.08 & 43.74 & 43.56 & $-1.2\%$ \\
fib / learned $W{=}8$  & 43.70 & 43.34 & 43.15 & $-1.3\%$ \\
pow2 / fixed $W{=}8$   & 44.78 & 44.98 & 49.86 & $+11.3\%$ \\
\bottomrule
\end{tabular}
\caption{FineWeb-Edu perplexity as a function of evaluation length for
models trained at length 1024. Dense is best at 1024 but collapses past it;
structured sparse attention is robust to 4$\times$.}
\label{tab:extrap}
\end{table}

\paragraph{Mechanism.}
Fixed-offset sparse attention only ever queries a bounded set of relative
positions, nearly all of which appear during training. RoPE rotation angles for
those offsets are therefore in-distribution at any sequence length. Dense
attention, by contrast, attends every relative position, including distances
between 1024 and 4096 whose rotary angles were never trained; vanilla RoPE
without scaling does not extrapolate to them, and perplexity blows up. The same
mechanism explains the ordering among sparse variants. Plain power-of-2 degrades
more ($+11.3\%$) because its reach-matched base introduces new large rungs (1024,
2048) at long evaluation length that were never trained; the Fibonacci base, with
a fixed 15-rung set, introduces no such new offsets. The coprime variant degrades
mildly ($+2.8\%$), making the linear stagger the more length-robust of the two.
The practical reading: at training length sparse attention pays a quality cost,
but it extrapolates at near-zero cost to lengths where dense attention is
unusable.

\subsection{Honest negatives}
\label{sec:negatives}

\paragraph{Training-length gap to dense.}
At length 1024 the dense baseline reaches 33.37 FineWeb perplexity versus 42.02
for the best sparse model, about 26\% better. Every sparse variant in the matrix
is well above dense at training length. At this scale, sparse attention is a
substantial quality sacrifice in-distribution; its value here is length
generalization, not in-distribution quality.

\paragraph{The gain is not long-range-specific.}
We resolve loss by token position within the 1024 window
(Table~\ref{tab:posloss}). If the staggering advantage came from wider long-range
coverage, the gap to fixed attention should grow at later positions, where more
distant context is available to reach. It does not: the staggered-minus-fixed gap
on FineWeb is roughly flat at $+0.035$ nats across all but the first position
bin. The perplexity improvement is a uniform, broadly distributed gain in
modeling quality, not a long-range-reaching effect. The effective-receptive-field
argument explains \emph{why} sparse attention works at all; it does not explain
\emph{where} the staggering gain comes from, which appears to be general
per-layer representational diversity.

\begin{table}[htbp]
\centering
\small
\begin{tabular}{lcccccccc}
\toprule
schedule & b0 & b1 & b2 & b3 & b4 & b5 & b6 & b7 \\
\midrule
fib / fixed $W{=}8$    & 3.883 & 3.808 & 3.767 & 3.770 & 3.742 & 3.780 & 3.776 & 3.763 \\
fib / stagger $W{=}8$  & 3.861 & 3.769 & 3.733 & 3.731 & 3.709 & 3.737 & 3.739 & 3.726 \\
\addlinespace
gap (fixed $-$ stagger) & .023 & .039 & .033 & .040 & .033 & .044 & .037 & .037 \\
\bottomrule
\end{tabular}
\caption{Position-resolved FineWeb NLL (nats), early context (b0) to late
context (b7). The staggering advantage is flat across position bins, not growing
at long range.}
\label{tab:posloss}
\end{table}

\paragraph{Auxiliary evaluations are capacity-limited.}
Beyond perplexity we logged a battery of auxiliary evaluations for every model,
including the dense baseline: two synthetic multi-hop reasoning tasks (RULER
variable tracking and LEGO) and a set of cheap probes carried over from the
training harness (LAMBADA, associative recall, factual cloze, BLiMP, and an
in-context-learning score). At 60M parameters nearly all of these are at chance or
at the floor and do not separate the attention patterns. LEGO holds its $50\%$
chance rate for every model ($0.47$--$0.51$ across all runs); RULER variable
tracking is near chance at about $11\%$ ($0.10$--$0.13$); associative recall is
zero for every model; LAMBADA top-1 accuracy is $2$--$5\%$ and factual cloze below
$2\%$; the in-context-learning score is too variable across runs to interpret. The
one exception is BLiMP, on which every model scores about $89\%$ (well above its
$50\%$ chance), but the spread across configurations is under five points and the
task probes grammatical acceptability rather than long-range spacing, so it does
not discriminate either. We report the full battery for completeness and rest the
quality claims on perplexity and the extrapolation result.

\subsection{Robustness to seed}
\label{sec:seed}

The full matrix uses a single seed. To bound the seed noise and re-test the
smallest margins, we retrained the five $W{=}12$ configurations at additional
seeds---three seeds each for the four integer-offset (bool-mask) configs, two for
the learned config, which is far costlier to train. Table~\ref{tab:seed} reports
the per-config mean and standard deviation. The seed noise is small on FineWeb
(per-config std $0.06$--$0.18$, pooled $\approx 0.13$), so the three headline
margins clear it comfortably: staggered beats fixed by $1.10$ ($\sim 8\sigma$),
beats learned by $0.77$ ($\sim 6\sigma$), and beats power-of-2 staggered by
$0.76$ ($\sim 5\sigma$). The two near-ties are confirmed as ties: staggered
versus coprime is $0.10$, smaller than either config's std and with overlapping
seed ranges; power-of-2 staggered versus learned Fibonacci is $0.01$.

\begin{table}[htbp]
\centering
\small
\begin{tabular}{lcccc}
\toprule
config ($W{=}12$) & seed 42 & seed 43 & seed 44 & mean $\pm$ std \\
\midrule
fib / stagger        & 42.02 & 41.96 & 42.26 & $42.08 \pm 0.13$ \\
fib / coprime        & 42.09 & 42.03 & 42.43 & $42.18 \pm 0.18$ \\
pow2 / stagger       & 42.69 & 42.77 & 43.08 & $42.84 \pm 0.17$ \\
fib / learned        & 42.85 & 42.85 & ---   & $42.85 \pm 0.00$ \\
fib / fixed          & 43.12 & 43.16 & 43.27 & $43.18 \pm 0.06$ \\
\bottomrule
\end{tabular}
\caption{Seed replication at $W{=}12$ (FineWeb-Edu perplexity). Four
integer-offset configs at three seeds, the learned config at two. Per-config std
is $0.06$--$0.18$; the three headline margins exceed it by $5$--$8\times$, while
staggered-vs-coprime and pow2-staggered-vs-learned are within it (ties).}
\label{tab:seed}
\end{table}

\section{Related Work}
\label{sec:related}

\paragraph{Fixed and per-head sparse patterns.}
Strided and log-sparse patterns \citep{child2019sparse}, local-plus-global
windows \citep{beltagy2020longformer,zaheer2020bigbird}, and geometric dilation
mixed within layers \citep{ding2023longnet} widen the receptive field through
composition or across heads. PowerAttention \citep{powerattention2025} frames
coverage through the stack explicitly but uses the same pattern in every layer.
Fibottention \citep{fibottention2024} uses Fibonacci/Wythoff dilation that differs
across heads and is deterministic and non-adaptive. Our staggering differs by
varying a single spacing scalar \emph{per layer} and by isolating that choice in a
matched language-modeling comparison.

\paragraph{Per-layer spacing schedules.}
WaveNet \citep{oord2016wavenet} stacks per-layer increasing dilation for an
exponential receptive field at constant parameters; the gridding analysis of
\citet{wang2018understanding} shows that common-factor dilation rates leave
periodic gaps and that coprime rates avoid them. Our coprime variant ports that
fix to attention; we find it does not improve language-modeling perplexity over a
plain linear ramp at this scale, and is slightly less robust under length
extrapolation. DilateFormer \citep{jiao2023dilateformer} varies dilation per head
rather than per layer. Closest to our setting, MSWA \citep{xu2025mswa} staggers
the \emph{local window size} across both heads and layers, widening it from
shallow to deep layers for broader receptive field through the stack. The
distinction is the quantity being staggered: MSWA varies the contiguous window
\emph{width}, whereas we hold the local window fixed and stagger the
\emph{sparse Fibonacci dilation spacing}: the offsets at which each layer
samples beyond the window.

\paragraph{Learned spacing.}
Adaptive attention span \citep{sukhbaatar2019adaptive} learns a per-head/per-layer
scalar controlling a contiguous attention reach via a soft-mask cutoff; that
parameter trains cleanly. Dilated convolution with learnable spacings
\citep{khalfaoui2023dcls} learns spacing in convolutions. In contrast to the
span-cutoff case, our learned multiplicative dilation scalar is inert (it does
not move appreciably and does not beat a fixed schedule), which we attribute to a
flat loss landscape in the dilation multiplier.

\paragraph{Length extrapolation.}
ALiBi \citep{press2021alibi} biases attention by distance to extrapolate, and RoPE
\citep{su2021roformer} is the position scheme we use. Our extrapolation result is
not a new positional method but an observation: fixed-offset sparsity confines
attention to trained relative positions, so it inherits robust extrapolation from
RoPE at no extra cost, while dense attention with the same RoPE does not.

\section{Discussion and Limitations}

The headline practical finding is a trade: structured sparse attention costs
in-distribution quality at training length but extrapolates to several times that
length where a recipe-matched dense model with the same positional scheme fails.
The within-sparse design lessons are that a static per-layer stagger beats both a
fixed schedule and a learned one, that the effect is base-agnostic, and that the
Fibonacci base adds a separable gain.

Several limitations bound these claims. The full matrix uses a single seed; we
replicated the five $W{=}12$ configurations at two to three seeds
(Section~\ref{sec:seed}) and find the three headline margins exceed the seed
noise by $5$--$8\times$, while the two near-ties are within it. The remaining
windows ($W{=}6,8,10$) are still single-seed, and the consistency of the
learned-versus-staggered and staggering-versus-fixed gaps across windows and
datasets is supporting rather than independent evidence at those windows. The
scale is 60M parameters, where the reasoning
tasks are at chance and the absolute perplexities are high; whether the
extrapolation advantage and the staggering gain persist at larger scale is open.
The extrapolation result is specific to vanilla RoPE without scaling; length
extrapolation methods built for dense attention would narrow or close the dense
collapse, and the fair comparison at that point is an open question. Finally, the
staggering schedule (a linear ramp over $[0.5,1.5]$) was fixed by design and only
lightly varied (the coprime reassignment); the optimal per-layer schedule is not
characterized here.

\section{Conclusion}

Across 21 matched language models, a static per-layer spacing stagger is the most
effective way to set sparse-attention offsets we tested: it beats a fixed
schedule, beats a learned per-layer scalar, and the gain transfers to a different
offset base. The mechanism we originally set out to study, a learned spring, is
inert. The result with the widest implication is that fixed-offset sparse
attention extrapolates to 4$\times$ training length at near-zero cost while dense
attention under the same positional scheme collapses, because sparsity confines
attention to relative positions seen during training. We report alongside these
the honest negatives that bound them: a training-length quality gap to dense and a
position-resolved analysis showing the staggering gain is broad rather than
long-range-specific.

\section*{Use of Generative AI}

The author used a generative AI language model (Anthropic's Claude) to assist
with drafting and editing the manuscript and with writing the training,
evaluation, and analysis scripts. All AI-assisted output, including every claim,
numerical result, and reference, was reviewed and verified by the author, who
takes full responsibility for the content of this paper.

\bibliographystyle{plainnat}
\bibliography{references}

\end{document}